\documentclass{article}
\usepackage{spconf,amsmath,graphicx}
\usepackage{times}
\usepackage{epsfig}
\usepackage{amssymb}
\usepackage{hyperref}
\usepackage{fancyhdr, lipsum}

\fancypagestyle{firstpage}{
   \fancyhf{} 
   \newcommand{\changefont}{%
    \fontsize{9}{11}\selectfont
    }
    \headheight 40pt

   \cfoot{\thepage}
   \fancyhead[L]{\changefont © 20XX IEEE.  Personal use of this material is permitted.  Permission from IEEE must be obtained for all other uses, in any current or future media, including reprinting/republishing this material for advertising or promotional purposes, creating new collective works, for resale or redistribution to servers or lists, or reuse of any copyrighted component of this work in other works.}
}
\title{Camera Calibration with Pose Guidance}
\name{Yuzhuo Ren, Feng Hu}
\address{NVIDIA, Autonomous Vehicle, Santa Clara, CA, USA}
\begin{document}
%
\maketitle
\thispagestyle{firstpage}
\begin{abstract}
Camera calibration plays a critical role in various computer vision tasks such as autonomous driving or augmented reality. Widely used camera calibration tools utilize plane pattern based methodology, such as using a chessboard or AprilTag board, user's calibration expertise level significantly affects calibration accuracy and consistency when without clear instruction. Furthermore, calibration is a recurring task that has to be
performed each time the camera is changed or moved. It's also a great burden to calibrate huge amounts of cameras such as Driver
Monitoring System (DMS) cameras in a production line with millions of vehicles. To resolve above issues, we propose a calibration system called Calibration with Pose Guidance to improve calibration accuracy, reduce calibration variance among different users or different trials
of the same person. Experiment result shows that our proposed method achieves more accurate and consistent calibration than traditional calibration tools.
\end{abstract}
\section{Introduction}
\label{sec:intro}
Camera calibration models and estimates a camera's intrinsic and extrinsic parameters, and is an essential first step for many robotic and computer vision applications~\cite{liu2017calibration,sochor2017traffic,wang2017calibration, ren2013techniques}. Intrinsic parameters deal with the camera's internal characteristics, such as, its focal length, principle point, skew, and lens distortion~\cite{zhang2000flexible}. Extrinsic parameters describe camera's position and orientation~\cite{huang2019research,ramirez2020improve}. Knowing a camera's calibration parameters allows us to remove its lens distortion, which is necessary in many applications that demands accuracy such as vehicle or pedestrian detection in wide Field of View autonomous vehicle cameras. However, reliable and accurate camera calibration usually requires an expert intuition to reliably constrain all of the parameters in the camera model. Existing calibration toolboxes~\cite{bradski2000opencv, bouguet2004camera} ask users to capture images from a posed calibration pattern board (chessboard~\cite{zhang2000flexible, chen2019plane, li2011geometric}, circle grid pattern~\cite{ha2017deltille}, AprilTag~\cite{xie2018infrastructure, sagitov2017artag}, etc.) in positions of their choosing, after which the maximum-likelihood calibrations parameters are computed using all images in a batch optimization. Tan \textit{et al.}~\cite{tan2017automatic} proposed to use monitor to display poses, however how to choose optimal poses to display on monitor screen is not considered. Richardson \textit{et al.}~\cite{richardson2013aprilcal} and Rojtberg \textit{et al}.~\cite{rojtberg2018efficient} proposed pose selection for interactive calibration which depends on a good pose initialization. The existing calibrators have common issues: 1) Calibration result consistency is not guaranteed if a tool is ran by users with different level of expertise.  Even for the same user, different runs of the same tool may result in significant difference. 2) The widely used re-projection error alone is not sufficient to control estimated parameters' error. 3) User has to guess the chessboard pose and whether the pose number and variation can lead to a successful calibration, this is challenging especially for the novice without domain expertise. It can cause frustrated user experience and also make quality control hard~\cite{sturm1999plane}. 

\begin{figure}
\centering
\includegraphics[height=4cm]{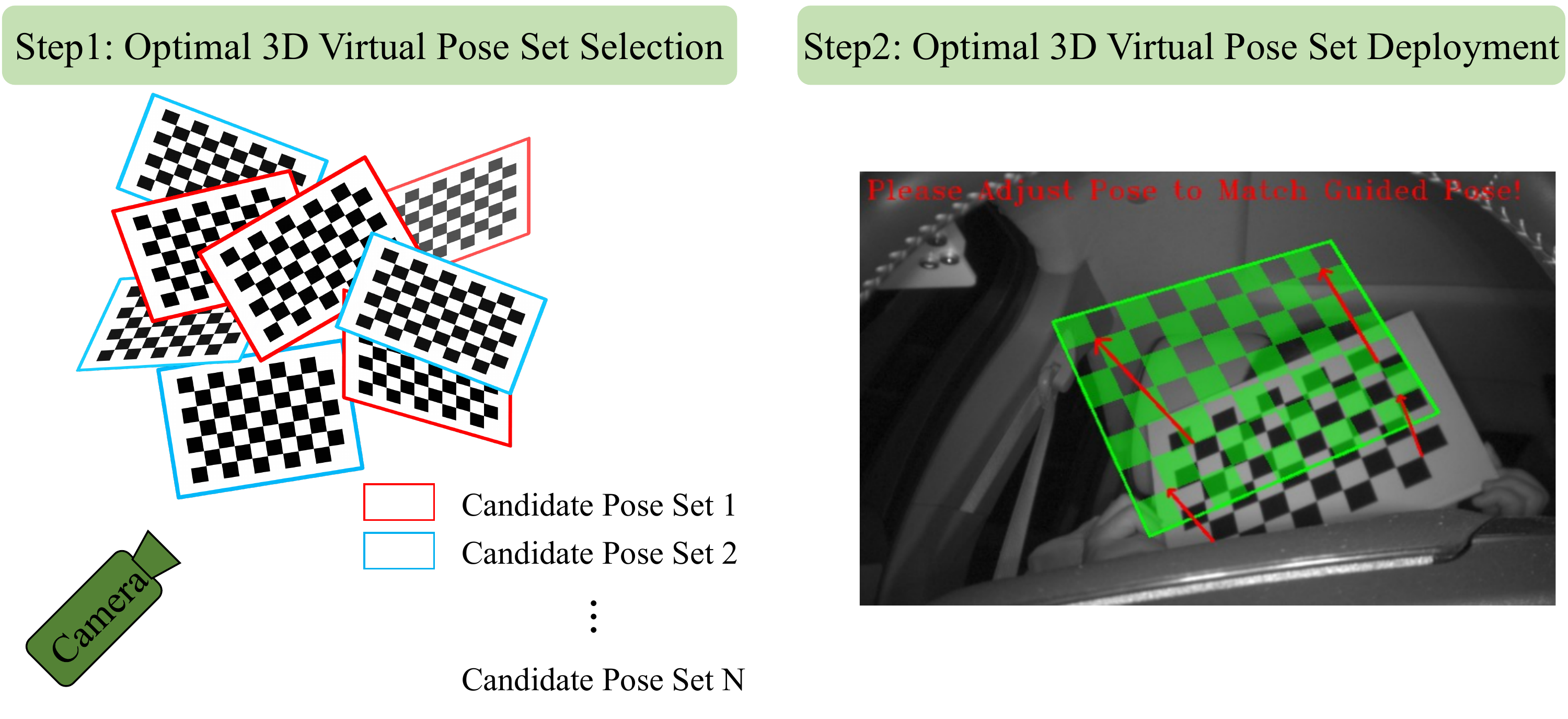}
\caption{Overview of proposed camera calibration with pose guidance system. In the first step, our  approach generates a set of 3D virtual poses. The optimal 3D virtual set is selected among many candidate pose sets. Two pose sets are denoted using red and blue color are shown for illustration. The pose set maximizing a defined score function is selected as optimal pose set which is deployed in our calibration system in next step. In the second step, an expected virtual pose is displayed on top of video streaming to guide the user to move calibration pattern with adjustment instructions shown in red arrows.}
\label{flow}
\end{figure}

In this work, we try to close these gaps by proposing the Calibration with Pose Guidance system, the diagram of which is shown in Fig.~\ref{flow}. In the first step, a set of optimal 3D virtual poses are selected using a novel score function which narrows down solution search space and avoid degenerated poses. In the second step, expected virtual poses are displayed on top of video streaming to guide the users to move calibration pattern respectively, with adjustment instructions shown in red arrows.

We summarize our major contributions as follows: 1) A method to automatically generate an optimal set of poses for calibrating a camera is proposed. The pose set automatically avoids degenerated cases, such as feeding the images captured at the same place many times into the tool and helps narrow down solution search space for calibration optimization. 2) A novel score function to evaluate pose sets to find the optimized set for a specific application scenario. 3) A gamified Human Computer Interface (HCI) that is simple and straightforward to guide any user, no matter of its expertise, to capture sufficient and desired pre-defined poses accurately and consistently, with visual hints for adjustment of the pattern board for each pose. Our method saves a lot of training time for novices to conduct calibration. The optimal pose set achieves higher accuracy and consistency than human involved calibration method especially for the novices.

\section{Methodology}
\label{sec:pagestyle}

\subsection{Calibration with Pose Guidance System Overview}
While the same methodology applies to all kinds of camera models and pattern boards, we use the pinhole camera model and chessboard~\cite{zhang2000flexible} to illustrate our proposed method. Denote an arbitrary 3D world point as $\bold{M} = {[X, Y, Z]}^{T}$ and its projected 2D image point $\bold{m} = {[u, v]}^{T}$, their homogeneous representations can be denoted as $\tilde{\bold{M}}= {[X, Y, Z,1]}^{T}$ and $\tilde{\bold{m}}= {[u, v,1]}^{T}$. Their geometric relationship can be represented as the following equation~\cite{zhang2000flexible},

\begin{equation}
s\tilde{\bold{m}} = \bold{K} \Delta([\bold{R}\:\bold{t}] \tilde{\bold{M}}), \\
\label{3d2drelation}
\end{equation}



where $[\bold{R}\:\bold{t}]$ are the rotation and translation which relate the world coordinate system to the camera coordinate system, $\bold{K}$ is the intrinsic matrix, $s$ is the scalar factor, and $\Delta (\cdot)$ is the distortion operator. We use $k_{1},k_{2},k_{3}$ to denote radial distortion, and $p_{1}, p_{2}$ to denote tangential distortion. For intrinsic matrix, we use $f_{x}, f_{y}$ to denote camera focal length, and $c_{x}, c_{y}$ to denote the principal point. Camera intrinsic calibration is to estimate $\bold{K}$ and lens distortion $\Delta$. Camera extrinsic calibration is to estimate $\bold{R}$ and $\bold{t}$. Intrinsic and extrinsic parameters can be estimated by optimization procedure ~\cite{zhang2000flexible} to minimize the re-projection error in Eq.(\ref{re-projectionerror}),


\begin{equation}
\epsilon_{\text{repoj}} = \sum_{i}^{N} \sum_{j}^{M} ||\bold{m}_{i,j}-\hat{\bold{m}} (\bold{K}, \bold{R}_{i}, \bold{t}_{i}, M_{i,j})||,
\label{re-projectionerror}
\end{equation}

where $\hat{\bold{m}} (\bold{K}, \bold{R}, \bold{t}, M)$ is the projection of point $M_{j}$ ($j=1,2,...,M$) in image $i$ ($i=1,2,...,N$) according to Eq.(\ref{3d2drelation}). $\bold{m}_{i,j}$  is the correspondent detected 2D point for point $j$ in image $i$.

Our pose guidance contains two steps as shown in Fig.\ref{flow}, optimal 3D virtual pose set selection, and pose set deployment, the details of which are described in Section ~\ref{poseselection} and ~\ref{deployment} respectively.


\subsection{Optimal 3D Virtual Pose Set Selection}\label{poseselection}

We define a pose $p_{i}$ as the chessboard's posture in 3D space, which can be parameterized as $p_{i}({R}_{i}, {t}_{i})$. A pose set P is a set of N such poses, where N is experimental variable, e.g. N = 20 in our experiments, which can be represented in Eq.(~\ref{posesetselection}):

\begin{equation}
\bold{P} = \{p_i (R_i, t_i)|i = 1,2,...,N\}. 
\label{posesetselection}
\end{equation}



In previous work, no constraints are set for how such a pose set shall be selected, and they can be randomly picked up. However, random pose set can have multiple issues. First, pose set contains degenerated case leads to singular solution in calibration optimization step. Second, the coverage of the poses may not be sufficient horizontally, vertically, or in terms of distance, rotation angle variance, which are critical in many applications, such as accurate distortion parameter estimation.

There are two steps in our proposed optimal 3D virtual pose set selection: 1) proposing candidate virtual pose sets; and 2) defining a score function and searching in the candidate sets that maximizes the score function. Finding the optimal solution for optimal pose set, denoted as $\bold{p}^{*}$, is difficult due to the infinite of the searching space. However, we can set reasonably constraints according to specific application and propose candidate sets, for example, removing all poses whose camera-board distance greater than 2 meters or whose yaw angle greater than a certain degree threshold for a DMS application. Then, we define a novel score function to rank pose set candidates and select the one with the highest score as the final result.

\textbf{Generation of High Quality Pose Set.} For each computer vision application where camera calibration is required, we can define a pose search space $\textbf{S}_{a}(\textbf{R}_{a}, \textbf{t}_{a})$.
We define a pose search space $\textbf{S}_{a}(\textbf{R}_{a}, \textbf{t}_{a})$ which is the camera working field of view space. Note that $\textbf{S}$ can vary among different camera use cases. For example, in DMS we are interested to ensure objects within 1 or 2 meters in the camera's field of view are imaged appropriately, but the interested range can reduce to 10 to 80 centimeters if we are calibrating a smartphone front camera. Assume $M$ poses are uniformly sampled in $\textbf{S}_{0}$. We randomly select $N$ poses to avoid degenerate poses, such as repetition of the same pose, or missing coverage of a corner. For example, two parallel poses only have different distance to camera lead ambiguity to focal length estimation.  Calibration degenerated cases will result in local minimal solution and have been studied by many research ~\cite{sturm1999plane, hammarstedt2005degenerate, sturm2000case, buchanan1988twisted, triggs1998autocalibration}.


\begin{figure}
\centering
\includegraphics[height=5cm]{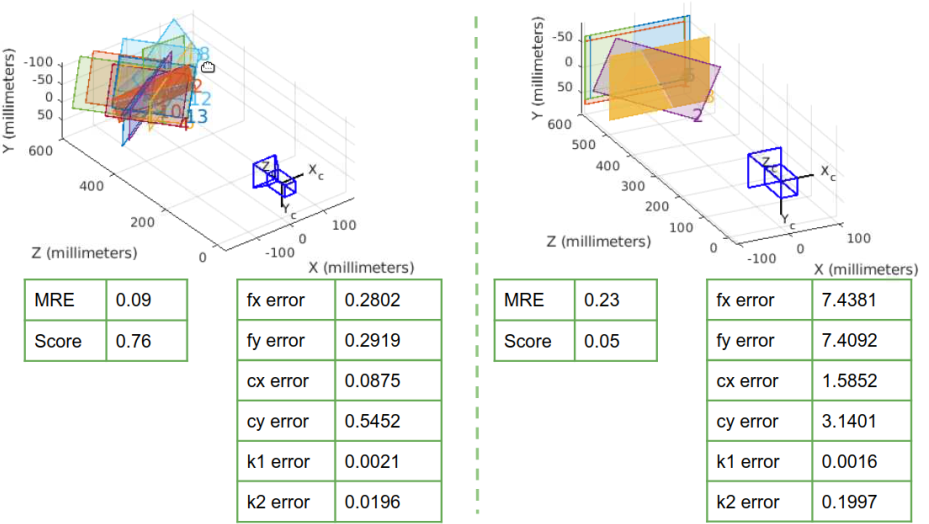}
\caption{Example of two pose sets evaluated using MRE and our proposed score function. The pose set covers the whole camera field of view evenly and consist of various pose variations gives higher score, and the bad pose set with many duplicate poses ranks low in our score. Both of their MRE are considered as good calibration if using industrial practical standard (i.e. MRE less than 1 or 2 pixels).}
\label{ranking}
\end{figure}
\textbf{Pose Set Ranking.} We adapt an iterative procedure to rank the pose set candidates~\cite{richardson2013aprilcal}: several poses are selected to estimate camera model, iterative the procedure to cover poses on not very well calibrated region and an updated camera model is estimated. We chose 15-20 (an empirical number from our experience and also suggested by many references\cite{bradski2000opencv}) selected poses cover whole camera field of view to estimate camera intrinsic parameters, denoted as $\bold{C}=[f_{x}, f_{y}, c_{x}, c_{y}, k_{1}, k_{2}, k_{3}, p_{1}, p_{2}]$. Note that the initialization step can be skipped if a good estimation of intrinsic parameters are already known, for example, camera factory calibration is known or the same model of camera has already been calibrated. Major previous work~\cite{zhang2000flexible} only use Mean Reprojection Error (MRE) to evaluate the quality of a calibration result. However, MRE alone is not sufficient for measuring all calibration intrinsic parameters' accuracy; MRE can still be very small for large intrinsic parameter errors, as shown in Fig.~\ref{ranking}. Our proposed score function takes both MRE and estimated parameter variance into consideration. Camera parameter is estimated from each pose set candidate, then we compare the camera parameter estimated from each pose set candidate with initialized camera model parameter $\bold{C}$. The score function to evaluate each pose set candidate's quality is reciprocal of summation of MRE and parameter estimation variance, as shown in Eq.(\ref{costfunction}),



\begin{equation}
S(\bold{P}|\bold{R}, \bold{t},\bold{K},\Delta(\cdot))=\frac{1}{\alpha\epsilon_{\text{repoj}} + \beta || \hat{\bold{C}} - \bold{C}||}, \\
\label{costfunction}
\end{equation}
where $\hat{\bold{C}}$ is the estimated intrinsic parameters, $\alpha$ and $\beta$ are parameters to control the cost from re-projection error and parameter estimation error.

The optimized pose set $\bold{P}^{*}$, which obtains the highest score among all the candidate sets is therefore defined as: 

\begin{equation}\label{equ:opt}
\textbf{P}^{*} = \operatornamewithlimits{argmax}\limits_{\bold{P}}S(\bold{P}|\bold{R}, \bold{t},\bold{K},\Delta(\cdot)).
\end{equation}

The score function is in favor of the pose set that gives both minimum re-projection and parameter estimation variance. The pose set with the highest score is chosen to be the final optimal pose set.

\begin{figure}
\centering
\includegraphics[height=6.5cm]{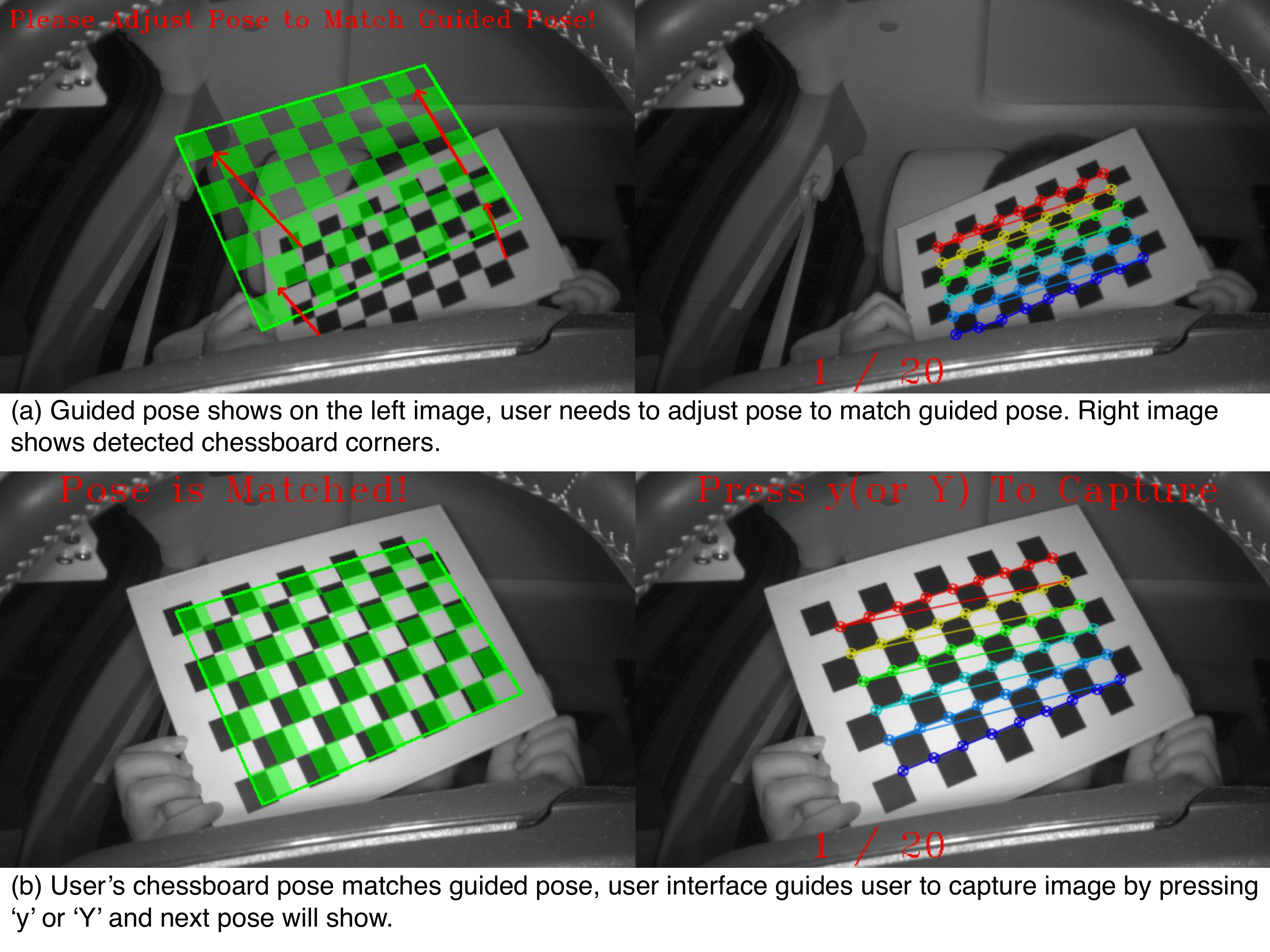}
\caption{Calibration with pose guidance system user interface.}
\label{semiautomatic}
\end{figure}

\subsection{Optimal 3D Virtual Pose Set Deployment}\label{deployment}
We discuss how to use optimal pose set to project onto calibration interface to guide a user to pose calibration pattern board appropriately in this section. One example of how a user is guided to move a chessboard, match the expected pose, and capture a qualified image is shown in Fig.\ref{semiautomatic}. To capture a qualified image, the user needs to move the calibration pattern around to ensure its image on screen matches the guided pose displayed. If the average distance of the four out-most corners is less than a threshold, where distance is defined as the $L_{2}$ pixel distance between the expected position and current position, the user' pose is considered matched with guided pose. Once the user matches the guided pose, the system will capture current frame and show the next guided pose. The procedure repeats until all $N$ images are captured. Our solution supports both automatically capturing, or manually capturing such as by pressing a specific key in keyboard.


\section{Experiments}
\label{sec:typestyle}
We evaluate our proposed camera calibration system from multiple perspectives. First, we evaluate our proposed score function to demonstrate its capability to select optimal pose set which improve calibration accuracy. Second, we report result to demonstrate the robustness of calibration accuracy and reproducibility of our method. Finally, we demonstrate that our calibration tool is applicable to a wide variety of lens\footnote[1]{Len1:HFOV=80, VFOV=60, resolution=1280x800, fomat=IR}
\footnote[2]{Len2:HFOV=120, VFOV=100, resolution=1920x1208, fomat=RGB}.


\subsection{Score Function Evaluation}
To evaluate the effectiveness of our optimal pose selection using defined score function in Eq.(\ref{equ:opt}), we simulate a virtual camera with known intrinsic parameters $\bold{C}$ and we use a $9\times6$ chessboard as calibration pattern. Pose selection space $\textbf{S}_{a}(\textbf{R}_{a}, \textbf{t}_{a})$ is chosen based on the working space range from specific use cases and candidate pose sets are generated in $\textbf{S}_{a}(\textbf{R}_{a},\textbf{t}_{a})$. Eq.(\ref{3d2drelation}) is used to project pose sets onto 2D image and Eq.(\ref{costfunction}) is used to compute score for each pose set.  

We show pose set with maximum and minimum Mean Re-projection Error (MRE), our proposed score and calibration parameter estimation error $||\hat{\bold{C}} - \bold{C}||$ in Table \ref{simulationresult}. Comparing first row and fourth row, pose set with smaller MRE may not have smaller parameter estimation error. In contrast to MRE, our score in favor pose set with both smaller MRE and smaller parameter variance which gives a much superior calibration accuracy evaluation. 

\begin{table}[h]
\begin{center}
\centering
\begin{tabular}{|c|c|c|c|c|c|c|c|c|}\hline
       & MRE    &  score in Eq.(\ref{costfunction}) &$||\hat{\bold{C}} - \bold{C}||$\\ \hline
PS(min MRE)    &0.0970 & 1.3689 &0.6336\\ \hline
PS(max MRE)    &0.1679 & 1.0820 &0.7562\\ \hline
PS(min score)  &0.1440 & 0.2052 &4.729\\ \hline
PS(max score)  &0.1186 & 4.8216 &0.088\\ \hline
\end{tabular}
\end{center}
\caption{Comparison of pose sets (PS) with maximum and minimum MRE, and with maximum and minimum score.}
\label{simulationresult}
\end{table}

\begin{figure}
\centering
\includegraphics[height=3.2cm]{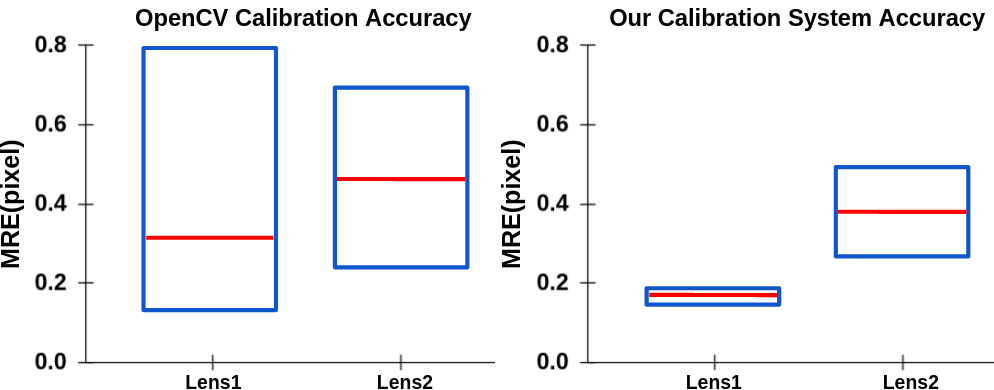}
\caption{Calibration accuracy and variance comparison between OpenCV and our calibration system. Mean re-projection error from 10 users' trials are indicated by red horizontal bars, the full range by blue boxes.}
\label{benchmark}
\end{figure}


\begin{table}
\centering
\begin{tabular}{lcccc}
\hline\noalign{\smallskip}
\bf{} & \multicolumn{2}{c}{\bf{OpenCV}}  &  \multicolumn{2}{c}{\bf{Ours}}\\

\noalign{\smallskip}
\hline
\noalign{\smallskip}
Len1& Mean & Std & Mean & Std\\
$f_{x}$ &  1357.8& 23.3&1350.1& \bf{2.6}\\
$f_{y}$  &  1356.9& 19.5&1352.1& \bf{2.8} \\
$c_{x}$ &  660.6& 14.2&657.6& \bf{2.7}   \\
$c_{y}$ &  411.4& 27.7 & 383.8& \bf{4.9}  \\
$k_{1}$ &  -0.3829 &0.0105 & -0.3774& \bf{0.0047}    \\
$k_{2}$ &  0.2436&  0.2715 & 0.2425& \bf{0.0472}  \\
\hline
Len2 & Mean & Std & Mean & Std\\
$f_{x}$ &  976.7& 13.8 & 972.5& \bf{2.9}\\
$f_{y}$ &  977.6& 13.4 & 974.4& \bf{2.9} \\
$c_{x}$ &  963.1& 9.5 & 954.9& \bf{4.2}   \\
$c_{y}$ &  633.8& 8.0 & 644.4& \bf{1.0}  \\
$k_{1}$ &  -0.3591& 0.0188 & -0.3454& \bf{0.0059}  \\
$k_{2}$ &  0.1777& 0.0375  & 0.1398& \bf{0.0092}  \\
\hline
\end{tabular}
\caption{Mean and standard deviation of focal lengths, focal centers and distortion parameters (only $k_{1}$ and $k_{2}$ are listed here for illustration) estimation for all trials in the human study. While the mean values of parameter estimation are similar from OpenCV and our method, our method provides much less standard deviation.}
\label{benchmark_2}
\end{table}


\subsection{Calibration Accuracy and Reproducibility}
We invited 10 participants without any previous calibration experience to calibrate various camera lens with two different methods: 1) OpenCV calibration toolbox~\cite{bradski2000opencv} 2) Calibration with Pose Guidance. Participants were given a printed instructions for calibration which describe change poses of chessboard when using OpenCV calibration. We show the participant sample poses from OpenCV calibration tutorial website. We received feedback like how much degree the chessboard pose should be changed, how far should be the distance between the chessboard and the camera, etc, which shows that participants in general need additional instructions to use OpenCV calibration tool.

Fig.\ref{benchmark} shows re-projection error using different calibration tools. Our proposed calibration system achieves much smaller re-projection error and smaller variance. Table \ref{benchmark_2} shows detailed statistics, where the mean and standard
deviation of estimated intrinsic parameter from 10 participants are listed. Our system provides much smaller standard deviations among all parameters estimation. In summary, our system provides: 1) smaller re-projection error which indicates higher calibration accuracy and 2) smaller parameter estimation variance among different trials which demonstrates stability and reproducibility of our calibration method.

\section{Conclusions}
Camera calibration with pose guidance is proposed to improve calibration accuracy, reduce calibration variance and reduce training time to novices. We propose a novel score function to select optimal pose set which reduces both re-projection error and intrinsic parameter estimation variance. Our proposed calibration system is evaluated against widely used calibration tools. Multiple experiments are conducted to demonstrate the accuracy and robustness of our system. 
\label{sec:majhead}
 
\bibliographystyle{IEEEbib}
\bibliography{egbib}
\end{document}